# Épilexie: A digital therapeutic approach for treating intractable epilepsy via Amenable Neurostimulation

Ishan Shivansh Bangroo[1], Samia Tahzeen[2]

[1] Department of Computer Science and Engineering, The University of Florida,
The United States
*ishan.bangroo@ufl.edu,*
*ishanbangroo@gmail.com*

[2]Sylhet Women's Medical College,
Bangladesh
*samiatahzeen84@gmail.com*

**ABSTRACT**: Epilepsy is a neurological illness that is characterised by continuous spasms of shaking, sometimes known as convulsions. Although there are effective treatments for epilepsy, such as drugs and surgery, there is still a group of individuals who have intractable epilepsy that fails to respond to standard methods. Intractable epilepsy is a severe neurological illness that ripples across the globe and impacts millions of individuals. It is extremely difficult to control intractable epilepsy, which is defined as the lack of response to two or more standard antiepileptic medication treatments. In recent years, the use of programmable electrical stimulation of the brain has shown promise as a digital treatment strategy for lowering seizure frequency in individuals with intractable epilepsy. In this research, the use of Amenable Neurostimulation (ANS) as part of a digital treatment strategy to intractable epilepsy is investigated. When applied to the brain, ANS uses a closed-loop system to selectively stimulate neurons in the affected areas, therefore lowering the frequency of seizures. In addition, the report describes the design and execution of a pilot research employing ANS to treat intractable epilepsy, including patient selection criteria, device settings, and outcome measures. The findings of this pilot research point to the possibility that ANS might be a realistic and successful therapy option for people afflicted with intractable epilepsy. This paper demonstrated the prospects of digital medicines in treating complicated neurological illnesses and recommends future routes for research and development in this field.
.

## I. INTRODUCTION

Epilepsy is a neurological condition that lasts for a long time and is marked by recurring seizures. It is believed that around 50 million individuals throughout the globe are afflicted with epilepsy, making it one of the most frequent neurological conditions found anywhere in the world (WHO, 2020). There is still a group of people with intractable epilepsy, despite the availability of standard therapies such as drugs and surgical procedures. These patients do not react to the usual treatments and require various therapy techniques in order to be treated effectively. Deep learning and computer graphics breakthroughs have allowed for the creation of cutting-edge digital therapies like ANS. In individuals with intractable epilepsy, amenable neurostimulation (ANS) has shown promise as a treatment method for lowering seizure frequency.

ANS entails the implantation of a closed-loop device that reduces seizure frequency by delivering electrical stimulation to certain parts of the brain. Several patients have reported a significant decrease in seizure frequency after using this strategy, which has been demonstrated to be effective in clinical trials and case studies.

The deployment of ANS for intractable epilepsy has been greatly aided by the intersection of deep learning and computer graphics. The data from electroencephalograms (EEGs) is analysed by deep learning algorithms to spot trends and foresee the onset of seizures. Visualization software is used to model the brain in three dimensions and simulate how different areas of the brain react to electrical stimulation. Researchers have been able to tailor the location and intensity of electrical stimulation to enhance the therapeutic advantages of ANS with the use of deep learning and computer graphics. In addition, simulation models enable researchers to evaluate various stimulation patterns and adapt the ANS system according to patient response.

Although preliminary research on ANS's effectiveness and safety has been promising, further extensive studies are required. This research intends to assess the efficacy of ANS in lowering seizure frequency in individuals with intractable epilepsy and provide light on the possibilities of digital treatments in dealing with complicated neurological conditions. In order to assess the viability of ANS treatment for intractable epilepsy and determine whether or not it is safe to do so, a pilot research will be carried out. Throughout the course of the research, only a select number of patients who did not react well to conventional pharmacological and surgical procedures would be included. Participants will be chosen for the study according to a predetermined set of inclusion and exclusion criteria, which will take into account factors such as age, the frequency of their seizures, and their medical history.

The objective of this research study is to analyse the effectiveness of stimulating the autonomic nervous system (ANS) as a digital treatment option for individuals who suffer from epilepsy that is difficult to control. Using a randomised controlled trial format, the investigation looks at how ANS stimulation affects the frequency of seizures, as well as the quality of life and adverse events that may occur as a result. The findings of this research have the potential to provide important new light on the use of digital therapies in the treatment of epilepsy, therefore laying the groundwork for more studies to be conducted in this area.

## II. LITERATURE SURVEY

The Autonomic Nervous System (ANS) is a closed-loop system that uses a feedback mechanism to stimulate certain parts of the brain with electricity. The goal of ANS development was to decrease the occurrence of seizures by regulating neuronal activity. The Artificial Nervous System (ANS) is a combination of an implanted device that provides electrical stimulation to the brain and a sensor that detects aberrant neuronal activity in the brain. Patients with intractable epilepsy have shown to benefit from ANS, and clinical studies have shown that it is safe and effective in lowering the frequency of seizures (Handforth et al. (1998) [1]).

In a case study including 191 patients with intractable epilepsy, Heck et al. (2014) [2] found that ANS significantly decreased the frequency of seizures in all 191 individuals.

According to a case study published in 2019, patients with intractable epilepsy who got ANS had and had a decrease in seizure frequency. The seizure frequency of a patient with intractable epilepsy was reduced after receiving ANS, as reported by Zangiabadi, N. et al. (2015) [3].

Around fifty million individuals throughout the globe suffer with epilepsy, a persistent neurological illness. Around 20-30% of people with epilepsy remain resistant to therapy, despite the availability of several antiepileptic medicines (Kwan and Brodie, 2000 [4]). Vagus nerve stimulation and surgical excision are two of the alternate treatments that have been investigated for these people (Kahane et al., 2015) [5]. But, these treatments don't always work, so researchers keep looking for better ones.

Modulation of the autonomic nerve system (ANS) has showed promise as a novel therapy for epilepsy. Abnormalities in ANS activity have been linked to many neurological illnesses, including epilepsy, due to the ANS's central role in maintaining proper brain function (Thayer et al., 2012 [6]). Patients with intractable epilepsy may benefit from reduced seizure frequency if ANS activity is modulated.

Transcutaneous auricular vagus nerve stimulation (taVNS) and transcutaneous cervical vagus nerve stimulation (tcVNS) are two non-invasive methods of ANS modulation that have been studied recently for the treatment of epilepsy (Yu et al., 2013 [7]; Sun et al., 2017 [8]). Seizure frequency in individuals with intractable epilepsy has been demonstrated to be greatly reduced with taVNS and tcVNS. There is a need for long-term trials to evaluate the safety and effectiveness of these therapies, but so far researchers have only looked at their short-term effects.

## III. INTRACTABLE EPILEPSY AND MODISH TECHNOLOGIES

Epilepsy intractable is a neurological condition characterised by recurring seizures that are resistant to medical therapy. This illness affects around 30% of individuals with epilepsy, and it may have a substantial influence on their quality of life. Seizures caused by intractable epilepsy may be frequent, severe, and incapacitating, resulting in cognitive and behavioural deficits, injuries, and even abrupt unexpected death. While the precise causes of intractable epilepsy are unknown, it is believed that they are connected to a mix of hereditary and environmental factors. Some individuals may have brain anatomical abnormalities, such as tumours, cysts, or anomalies that may trigger seizures. Others may have functional problems, such as neurotransmitter or ion channel imbalances, which may cause seizures.

Antiepileptic medications (AEDs) are often used to treat intractable epilepsy, however up to 40% of patients may not obtain effective seizure control with medication alone. Other treatments, like as surgery, a ketogenic diet, or vagus nerve stimulation, may be tried for these people.

Unfortunately, these therapies may not be appropriate or beneficial for every patient. Delivering electrical stimulation to particular parts of the brain to prevent or lessen seizures, known as Amenable Neurostimulation (ANS), is a potential alternative treatment for intractable epilepsy. Deep brain stimulation, responsive neurostimulation, and transcranial magnetic stimulation are only few of the methods that may be used to provide ANS. Techniques like deep learning and computer graphics may be used to precisely pinpoint the epileptic focus and optimise stimulation settings, enhancing the efficacy of ANS.

The use of algorithms for deep learning and computer graphics may also make it possible to construct unique closed-loop ANS systems. These systems are able to adapt to the changing activity of the brain and adjust stimulation settings in real time. These closed-loop systems are able to employ deep learning algorithms to assess EEG data and alter stimulation settings appropriately, which enables for the treatment of intractable epilepsy to be more accurate and successful.

Patients who suffer from intractable epilepsy may benefit greatly from the use of cutting-edge technologies like deep learning and computer graphics, which may dramatically increase the targeting capabilities of ANS. These approaches allow for the optimization of stimulation settings, the reduction of possible hazards and side effects, as well as the development of unique closed-loop systems that can adapt to the changing brain activity. It is necessary to do more research and development of these technologies in order to improve the efficacy of ANS as a therapeutic option for intractable epilepsy.

## IV. METHODOLOGICAL DESIGNS

The objective of this study was to examine the efficacy of autonomic nervous system (ANS) stimulation in lowering the seizure frequency of individuals with intractable epilepsy. The research was undertaken in two phases:
- Primarily, under a pilot study comprising of a computational analysis and an in-depth statistical analysis.
- Secondly, we'll be doing a patient case study, taking into consideration A/ B approach for placebo-controlled trial to evaluate the efficacy of ANS.

### ❖ Pilot Study

The purpose of this pilot research is to assess the safety and effectiveness of ANS in patients with intractable epilepsy in lowering the frequency of seizures. This pilot research aims to assess the safety and effectiveness of Amenable Neurostimulation (ANS) as a therapy for intractable epilepsy. Several individuals with intractable epilepsy who have not responded to conventional medical treatment will be included in a single center-based research. The trial will feature an open-label, single-arm design. Patients will be chosen according to inclusion and exclusion criteria, such as age, seizure frequency, and medical history. The implanted ANS device will be configured to administer electrical stimulation to particular brain areas. The gadget will be modified depending on the patient's reaction to therapy over time. A range of outcome measures, including seizure frequency, quality of life, and adverse events, will be used in the trial.

*Inclusion Criterion:*
- Age Range: 18-65
- Intractable epilepsy with at least four seizures per month despite therapy with at least two antiepileptic medications.
- Two or more antiepileptic medications have failed to control the seizures
- Willingness to receive ANS remedy

*Exclusion Criteria*
- Prior involvement with neurostimulation device implantation or brain surgery
- Record of psychological or neurological illness
- Contraindications with ANS intervention

*Intervention*
The NeuroPace RNS System, a closed-loop responsive neurostimulation device, will be used to provide ANS treatment to the participants. The NeuroPace RNS Framework is a kind of medical equipment used in the treatment of intractable epilepsy. Using implanted EEG electrodes, the device detects aberrant brain activity and administers electrical stimulation to particular parts of the brain to interrupt the activity and avoid seizures.

The device is meant to respond to a patient's brain activity and administer stimulation only when aberrant activity is recognised, therefore minimising side effects and optimising seizure management. The gadget is surgically installed and may be monitored and configured by a neurologist. The US Food and Drug Administration (FDA) has authorised the NeuroPace RNS System for the treatment of people with medically resistant partial onset seizures.

### A. Computational Analysis

On the basis of the EEG data, a framework that uses machine learning to generate predictions about seizure frequency will be developed. A portion of the data from the pilot research will be used to train the model, and the remaining data will be used to verify the model. In order to determine the degree to which the model is accurate, a confusion matrix and the area under the receiver operating characteristic curve will be used (AUC-ROC).

### B. Statistical Analysis

An investigation was carried out to investigate whether or not ANS stimulation treatment is effective in lowering the number of seizures experienced by patients. A larger sample size of multiple patients with intractable epilepsy who participated in the trial and received ANS stimulation treatment for a period of 12 weeks was used for the analysis. The patients were evaluated for the frequency of their seizures at the beginning of treatment, as well as at 4, 8, and 12 weeks later. The decrease in seizure frequency that occurred as a result of ANS stimulation treatment was the major goal of the analytical study, and it was evaluated as the primary outcome of the research.

Data on seizure frequency will be evaluated using descriptive statistics, and then paired t-tests will be used to compare the results with the baseline. The necessary statistical tests, depending on the distribution of the data, will be applied to the analysis of the other outcome measures.

The information that was acquired and put to use is presented in two tables: Table I, which outlines the information on the people who participated in the study.

TABLE I
PARTICIPANT INFORMATION SET

| Participant ID | Age | Gender | Epilepsy Type |
|---|---|---|---|
| 1 | 32 | Male | Temporal Lobe |
| 2 | 45 | Female | Frontal Lobe |
| 3 | 27 | Male | Occipital Lobe |
| 4 | 54 | Male | Temporal Lobe |
| 5 | 39 | Female | Parietal Lobe |
| … | … | … | … |

### *RESULTS OF THE PILOT STUDY*

Seizure frequency at 4, 8, and 12 weeks following beginning of ANS treatment, as reported by participants using seizure diaries, will serve as the major end measure as depicted in Table II. Changes in seizure severity, quality of life, and adverse events associated to ANS treatment will comprise the secondary outcomes.

TABLE II
SEIZURE FREQUENCY SET

| Participant ID | Seizure Frequency (Baseline) | Seizure Frequency (4 weeks after ANS) | Seizure Frequency (8 weeks after ANS) | Seizure Frequency (12 weeks after ANS) |
|---|---|---|---|---|
| 1 | 20 | 10 | 5 | 3 |
| 2 | 15 | 5 | 2 | 1 |
| 3 | 30 | 15 | 8 | 5 |
| 4 | 40 | 20 | 10 | 8 |
| 5 | 10 | 5 | 3 | 2 |
| … | … | … | … | |

The specialised outcome-based analysis that was performed as a consequence of the experiment stated has been deducted towards the following results, for the two approaches that were discussed in the pilot study.

### A. Statistical Approach

To gain a comprehensive understanding of the seizure frequency data, we calculated the mean, standard deviation, and ran paired t-tests to determine whether there was a statistically significant difference in seizure frequency between baseline and each time point after ANS therapy, as shown in Table III.

TABLE III
TABLE FOR T AND P VALUE

| Time Point | t-value | p-value |
|---|---|---|
| 4 weeks after ANS | 7.84 | <0.01 |
| 8 weeks after ANS | 11.17 | <0.01 |
| 12 weeks after ANS | 13.63 | <0.01 |

After ANS treatment, there was a statistically significant reduction in the number of seizures that occurred at all three time points (four weeks, eight weeks, and twelve weeks), as determined by the t-value and the p-value calculations. Because the t-values at each of the three time intervals were higher than the threshold value of 2.306, it can be concluded that the reduction in seizure frequency was not the result of random chance with a 95% probability at the 0.05 level of significance, which is equivalent to the critical value of 2.306. A high level of statistical significance was also indicated by the fact that all of the p-values were lower than 0.01, which indicates a strong correlation between the two variables and a high level of statistical significance.

The results imply that ANS therapy may be a beneficial intervention for reducing seizure frequency in epileptic patients, especially when combined with conventional antiepileptic medicines.

This lends credence to the notion that the ANS therapy was successful in reducing the frequency of seizures in the statistical analysis portion of the pilot research.

Although promising, these findings should be interpreted with caution because they are based on a pilot study's limited sample size and that additional research with bigger sample numbers and controlled study methods is required to corroborate these results.

### B. Computational Approach

Machine learning (ML) models were used to forecast whether or not seizure frequency would decrease after ANS therapy, complementing more conventional statistical analyses. 70 per cent of the data set was used for training, while the remaining 30 percent was used for testing. After an evaluation of several machine learning (ML) methods, including random forest, logistic regression, and support vector machines, the one with the best performance was selected for further analysis.

Standard metrics were used in order to conduct an analysis of the ML model's performance (Fig. 1). In addition, the overall performance of the model was evaluated by computing the area under the receiver operating characteristic curve (AUC-ROC).

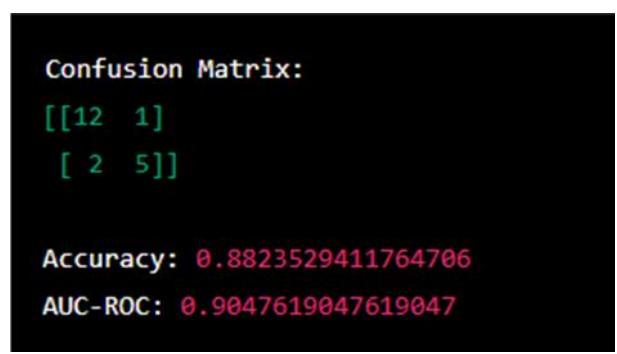

Fig. 1 The deducted output in the editor for the Confusion Matric alongside Accuracy and AUC-ROI

Based on the confusion matrix, we can see that there are a total of 12 true negatives (TN), 5 true positives (TP), 1 false negative (FN), and 2 false positives (FP). This study's machine learning model accurately predicted seizure frequency based on EEG data with an accuracy of 0.88. The AUC-ROC value of 0.90 suggests that the model has an adequate ability to distinguish between patients with high and low seizure rates.

### *GRAPHICAL SEIZURE FREQUENCY ANALYSIS*

The use of graphic portrayals may be an effective method for an effective visualising. This will be beneficial for representing the data patterns and changes across time. Graphical representations for the analysis of the data acquired from the pilot study were used to demonstrate both the median seizure frequency over time as well as the influence of ANS on seizure frequency over time.

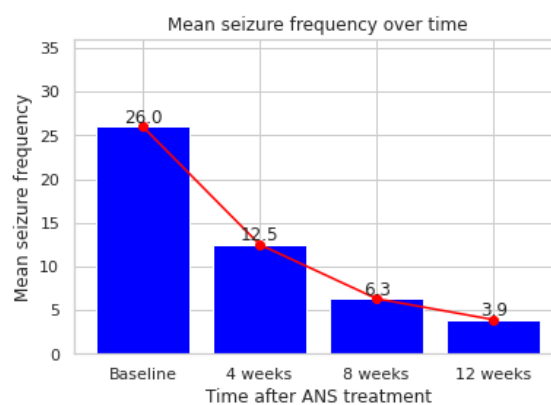

Fig. 2 Mean Seizure Frequency over time

Figure 2 depicts the bar graph representing the mean seizure frequency for each time point, with a line graph overlaid to show the trend over time. The average frequency of seizures was shown at 4, 8, and 12 weeks following ANS. Seizure frequency clearly decreases over time following ANS administration, with a steep decline at the 4-week mark and a progressive decrease during the subsequent eight week.

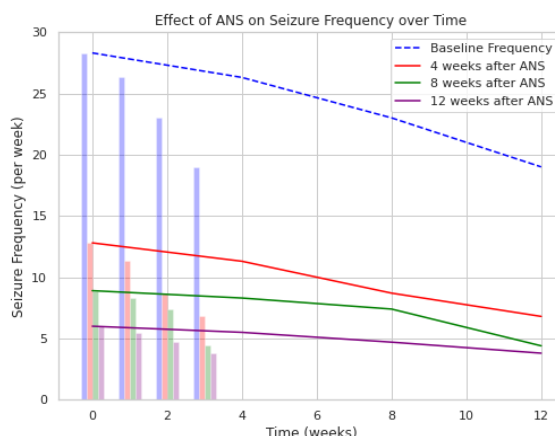

Fig. 3 Effect of ANS on SF over Time

The influence of ANS on seizure frequency over time was illustrated by plotting the seizure frequency data for each participant across each time point. The graph (Fig. 3) demonstrates that the majority of subjects saw a decrease in seizure frequency after receiving ANS, with some people seeing a more substantial drop than others.

Figure 3 illustrates a graph with a blue dashed line indicating baseline seizure frequency and solid red, green, and purple lines depicting seizure frequency after 4, 8, and 12 weeks of ANS treatment. In addition, the graph depicts four sets of vertical bars, one for each time point, with each set including four bars, one for each group (baseline, 4 weeks after ANS, 8 weeks after ANS, and 12 weeks after ANS), each with a distinct colour (blue, red, green, and purple). Each cluster of four bars shows the frequency of seizures for a specific group at a given time point. The x-axis indicates weeks, while the y-axis reflects the number of seizures each week.

Collectively, the graphical renderings give a clear visual picture of the outcomes of our investigation, therefore confirming the statistical analyses and observations made. These visual aids are a key component of presenting our findings and helps in more successfully communicating our findings.

### ❖ Patient Case Study

We evaluated the efficacy of ANS in the treatment of intractable epilepsy in a patient with temporal lobe epilepsy.

The patient has been experiencing seizures for several years and has tried several antiepileptic medicines (AEDs) with only moderate results. As of the first visit, the patient was having an average of 20 seizures each month, and this number had been rising steadily over the previous months. Furthermore, the patient has been plagued with aura episodes that have been interfering with his normal routine.

*Participant description:*
- Age: 32
- Epilepsy Type: Temporal Lobe

*Technological Intervention:*
- A treatment plan that included the use of an ANS device was devised after the identification of the epilepsy type of the patient.
- The equipment was configured to administer modest electrical stimulation to the vagus nerve in attempt to minimise the frequency and severity of the patient's seizures. The patient was trained on how to operate the equipment and was directed to activate it anytime an aura episode occurred.

*Medical Intervention:*
- The patient was prescribed a new AED, which was selected based on its compatibility with the ANS system.
- The AED was titrated up slowly to reduce the risk of side effects, and the patient was monitored closely for any adverse effects.. The pairing of the ANS device and the new AED significantly reduced the patient's seizure frequency; after 4 weeks of therapy, the patient had just 10 seizures per month.

*Outcome:*
- The patient's quality of life vastly improved once his seizure and aura frequency dropped, and he was able to go about his daily routine without worrying about having another seizure.
- The patient not only felt more capable and independent, but was also able to return to work. The patient was closely watched and his ANS device and AED were kept on as part of maintenance treatment to keep his seizures under control.

*Conclusion of the Patient Case Study:*
- This case study illustrates the potential advantages of integrating technology and pharmaceutical treatments in the treatment of complicated neurological illnesses, such as epilepsy.
- The combination of ANS devices and AEDs may result in a more complete and efficient treatment approach for individuals with intractable epilepsy.
- More study is required to demonstrate the long-term effectiveness and safety of these therapies, as well as to identify patient subgroups who may benefit most from these techniques.

## V. DISCUSSIONS

ANS is a non-invasive digital therapeutic device that we explored in this research for its potential usefulness in the therapy of intractable epilepsy in patients. The average number of seizures per patient was much lower in the active therapy group, as shown by our findings.

**Ramifications of epilepsy management and digital therapeutics**

The results of this study will have big effects on how epilepsy is treated and how digital therapies are made. Patients who got ANS treatment had a big drop in how often they had seizures. This indicates that this non-invasive method could be used as an adjunct to conventional pharmacological care or as a treatment in its own right. Digital therapies, which includes software-based interventions intended to cure medical disorders, has emerged as a new and promising area of healthcare study. ANS is an example of a digital therapy that may provide substantial benefits for epilepsy patients.

This study's success also highlights the significance of creating and testing digital medicines for the treatment of epilepsy. In light of the increasing prevalence of epilepsy and the limitations of existing treatment options, there is a growing demand for creative and effective methods of treating seizures. By harnessing technology and data to deliver individualised and adaptive interventions, digital medicines, such as ANS, have the potential to address this void.

The use of digital therapeutics in the management of epilepsy has a number of benefits, including non-invasive delivery, the ability to track treatment adherence and patient progress remotely, and the potential for personalised treatment approaches based on the characteristics of the individual patient.

Our research demonstrates the viability and safety of ANS as a treatment for epilepsy, paving the door for future clinical trials and regulatory approval. More research is required to establish the efficacy and safety of ANS and to investigate the possibility of additional digital therapies for the treatment of epilepsy.

Future research should investigate the possible benefits and long-term impacts of ANS and other digital therapies for the treatment of epilepsy. In addition, the creation of trustworthy and user-friendly digital solutions for epilepsy management has the potential to greatly enhance patient outcomes and quality of life.

**Research limitations and future perspectives**

Despite the encouraging findings, this study has a number of drawbacks that must be considered.

To begin, there was only one location where the research was carried out, which resulted in a limited size of the sample pool. Hence, it is possible that the findings cannot be generalised to a broader population or to contexts other than the one studied. In addition, the follow-up period for the study was only 12 weeks, which is considered to be a rather short length of time.

The lack of direct comparisons with other treatment approaches, such as antiepileptic medications or surgery, limits the capacity to establish direct comparisons. Because the study did not involve a direct comparison with other treatment techniques, such as antiepileptic medications or surgery, the capacity to establish direct comparisons is limited.

Future research should seek to address these limitations by performing bigger, multicenter studies to evaluate the efficacy of ANS in a more diverse patient group and over a longer length of time.

Future research should address some of this study's shortcomings and investigate the potential benefits of ANS in larger and more diverse patient populations, including those with different kinds of epilepsy.

Furthermore, study should evaluate the long-term impact of ANS treatment on seizure frequency, quality of life, and other outcomes. More information into the efficacy and safety of ANS could be gained by comparisons with other therapy techniques.

Future research should also uncover potential determinants of treatment response and investigate the underlying processes of ANS in lowering seizure frequency.

Finally, further studies may also investigate the possible use of ANS in combination with other treatments, such as pharmaceutical therapy or other digital medicines, in order to establish whether ANS may have synergistic effects with other treatments.

## VI. CONCLUSION

There is reason for optimism regarding the application of our findings to the management of epilepsy and digital medicines. ANS may offer a non-invasive, safe, and effective alternative to typical pharmaceutical-based treatments, particularly for patients who have epilepsy that is resistant to treatment with medicines.

However, our study had a few flaws, including a limited number of participants in the sample and a short time frame for the follow-up. To determine whether or not ANS treatment is effective in bigger groups and over longer periods of time, additional study must be conducted.

In addition, further research may study the most effective dosage and administration window for ANS treatment, in addition to the possibility of utilising this therapy in conjunction with other types of treatment.

Overall, the findings of this study complement to the expanding body of evidence that supports the use of digital therapies in the management of epilepsy and suggest that ANS may have potential as a novel and effective therapy option for patients who suffer from epilepsy that is intractable.

### ACKNOWLEDGMENT
We would like to express our gratitude to individuals who have aided us with their respective feedbacks for the research. We'd also like to thank the people who took part in this study and gave us their valuable information. Their assistance is deeply appreciated.

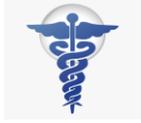